\documentclass[runningheads]{llncs}
\usepackage{comment}
 
\usepackage{eccv}

\usepackage{eccvabbrv}

\usepackage{graphicx}
\usepackage{booktabs}
\usepackage{wrapfig}
\usepackage[accsupp]{axessibility}  

\usepackage{hyperref}

\usepackage{orcidlink}
\usepackage{multirow}

\begin{document}

\title{Triangular Consistency as a Universal Constraint for Learning Optical Flow} 

\titlerunning{Triang. Consist. as Univ. Constraint for Learn. Optical Flow}

\author{Yi Xiao\inst{1} \and
Carlos Rodriguez Coronel\inst{1} \and
Jing Zhan\inst{1} \and
Haniyeh Ehsani Oskouie\inst{2} \and
Alex Wong\inst{3} \and
Dong Lao\inst{1}\\
Correspondence to: \email{\{yi.xiao,dong.lao\}@lsu.edu}}

\authorrunning{Y.~Xiao et al.}

\institute{Louisiana State University, Baton Rouge, LA 70803, USA \and
University of California, Los Angeles, Los Angeles, CA 90025, USA \and
Yale University, New Haven, CT 06520, USA
}

\maketitle

\begin{abstract}
We propose triangular consistency as a first-principled constraint for optical flow, which is agnostic to network architecture, supervision type, and dataset, and applies to both image-pair and multi-frame settings. This simple but powerful constraint is to compose two flows to induce a third flow and enforce consistency among the three. The composed flows may arise from (i) image pairs, yielding cycle consistency; (ii) multiple video frames, producing longer-range motion through temporal chaining; or (iii) image pairs combined with controlled synthetic transformations, which becomes data augmentation. This triangular consistency introduces negligible computational overhead and requires no additional annotations. Since it is derived directly from the geometry of optical flow, it does not rely on model-specific assumptions and serves as a ``universal'' plug-and-play component for optical flow training. Experiments show consistent improvement across supervised, unsupervised, and transfer learning settings. Code: \url{https://github.com/lsuvision/tri-flow}.
\end{abstract}

\section{Introduction}
\label{sec:intro}
The concept of optical flow was introduced by J. J. Gibson in the 1940s to describe visual motion projected onto the retina~\cite{gibson1950perception}, decades before the emergence of computer vision. When the term was later adopted by computer scientists~\cite{horn1981determining}\footnote{Lucas-Kanade~\cite{lucas1981iterative} appeared slightly earlier; however, Horn-Schunck~\cite{horn1981determining} is typically credited with introducing the term \emph{optical flow} into the computer vision literature.}, our computational system inevitably converted this instantaneous quantity into a problem of inference from \emph{discrete} frames, usually frame pairs. At first glance, this appears to be a straightforward computational discretization. However, if we consider the underlying physical process, optical flow, at its core, measures a continuous non-rigid transformation of coordinate systems of the underlying scene. Different temporal samplings of the same process must therefore obey a fundamental physical rule: they agree under composition~\cite{lao2018extending,le2024dense,neoral2024mft}. If one mapping transports coordinates from an initial configuration to an intermediate one, and another transports them from the intermediate configuration to a final one, then their composition uniquely determines the transformation between the initial and final coordinates. Such transformations can be chained indefinitely, yet consistency always holds. As such, compositional consistency is \emph{first-principled}: it is neither a computational assumption nor a regularization heuristic; it is directly grounded in the physics of the scene.

This knowledge by itself is nothing new. Multi-frame optical flow~\cite{sand2008particle,garg2010dense,ricco2012dense} and point tracking~\cite{karaev2024cotracker,cho2024flowtrack,le2024dense} have long used chained optical flow to improve long-range estimation. However, this compositional nature is not made explicit as a regularizing constraint. Modern learning-based optical flow methods~\cite{dosovitskiy2015flownet,sun2018pwc,teed2020raft}, in contrast, are predominantly formulated as predicting displacement between two frames. While some multi-frame paradigms~\cite{janai2017slow,janai2018unsupervised,liu2020learning,shi2023videoflow,dong2024memflow} also exploit consistency among more than two frames, they mostly build upon linear motion (constant speed) assumptions. The constraint that optical flows should agree under composition remains largely overlooked.

In this paper, we take the first step to fill this gap by studying the simplest form of composition: we consider only three frames (i.e., \emph{triangular}): given two consecutive optical flows, their composition should match the flow directly estimated between the first and the last frames. Cycle consistency~\cite{ren2017unsupervised}, where forward and backward flows between image pairs result in the identity mapping, becomes a special case, while temporal chaining of three consecutive frames and consistency under asymmetric transformations arise from the same principle. This first-principled relation provides a \emph{universal} supervision signal: when ground truth is unavailable, it acts as a constraint for self-supervision; when ground truth is available, it enables controlled data augmentation by generating new pseudo-labels. To our knowledge, triangular consistency has not been systematically adopted by optical flow literature, and we present the first comprehensive evaluation of this principle across supervised, unsupervised, and transfer learning, spanning multiple architectures and datasets. Through the experiments, we examine how far compositional consistency can improve optical flow accuracy. Specifically, our contributions are:

\begin{itemize}
\item \textbf{A universal geometric constraint.} We formalize \emph{triangular consistency}, the minimal compositional relation among three flows, and instantiate it as a training loss agnostic to architectures, supervision types, and datasets.
\item \textbf{Three training regimes, one principle.} We show how the same compositional rule yields (i) temporal compositional supervision on frame triplets, (ii) cycle consistency between image pairs, and (iii) a data augmentation scheme generating pseudo-ground-truth optical flow analytically.
\item \textbf{Practical integration.} The resulting training losses are lightweight, require no additional labels, and can be added to existing pipelines without modifying the estimator.
\item \textbf{Consistent empirical gains.} Across transfer, unsupervised, and supervised training, triangular consistency improves accuracy and/or cross-dataset generalization, with strong gains in transfer and out-of-domain evaluation.
\end{itemize}

We observe up to 18.1\% gain under single-epoch adaptation, 6-8\% improvements under unsupervised training, and up to 23.1\% cross-dataset gain under supervised training.
In summary, triangular consistency provides a simple yet principled mechanism to improve optical flow learning. It can be readily integrated into existing training pipelines without architectural modifications or additional human supervision, yielding consistent improvements.

\section{Related Work}
\label{sec:related}

Optical flow literature is extensive, and we only highlight some of the most relevant advancements, especially those related to consistency.

\noindent\textbf{Optical flow} was formulated with explicit data terms and regularization in earlier model-based and variational literature~\cite{horn1981determining,bailer2015flow,chen2016full,sun2010secrets,brox2010large,chen2013large,revaud2015epicflow,yang2015coarse,xu2011motion,yang2017s2f}. Deep-learning-based methods largely inherit this framing but replace hand-crafted priors with learned matching and refinement by neural networks. Representative milestones include FlowNet~\cite{dosovitskiy2015flownet}, pyramidal warping and cost volumes in PWC-Net~\cite{sun2018pwc}, and the strong all-pairs iterative paradigm of RAFT~\cite{teed2020raft}, followed by architectural variants improving either efficiency or robustness~\cite{wang2024sea,huang2022flowformer,sui2022craft,zhao2022global,sun2022skflow,dong2023rethinking,yang2018conditional}. While these works mainly advance architectures and data scale, our focus is orthogonal: we introduce a first-principled \emph{supervision} derived from the compositional nature of displacement fields. Therefore, it can be integrated into existing training pipelines without modifying the estimator.

\noindent\textbf{Consistency as supervision signal.}
When ground-truth flow is unavailable, supervision is commonly derived from photometric consistency: the predicted flow warps the target image, and the discrepancy between the source image and the warped target image provides a supervision signal~\cite{yu2016back,ren2017unsupervised}. Because photometric constancy is imperfect and breaks under occlusion, later pipelines typically incorporate robust penalties, explicit occlusion reasoning, or distillation/pseudo-labeling to filter unreliable regions~\cite{wang2018occlusion,meister2018unflow,janai2018unsupervised,Liu_2019_CVPR,liu2019ddflow,jonschkowski2020matters}. A complementary family of supervision signals is motion consistency that enforces agreement among predictions produced from different directions, frames, or transformations~\cite{liu2020learning,Stone_2021_CVPR,jeong2022imposing}. Specifically, geometric augmentation has also been explored by augmenting the source and target image with the same (\emph{symmetric}) transformation (e.g. rotation), so that the optical flow can be augmented accordingly~\cite{liu2020learning}. These methods motivate our work, but typically instantiate specific proxy constraints tied to particular training setups. In contrast, we derive a single compositional rule as a universal constraint that subsumes cycle consistency while generalizing to temporal chaining and \emph{asymmetric} augmentation.

\begin{figure*}[t]
\centering
\includegraphics[width=\textwidth]{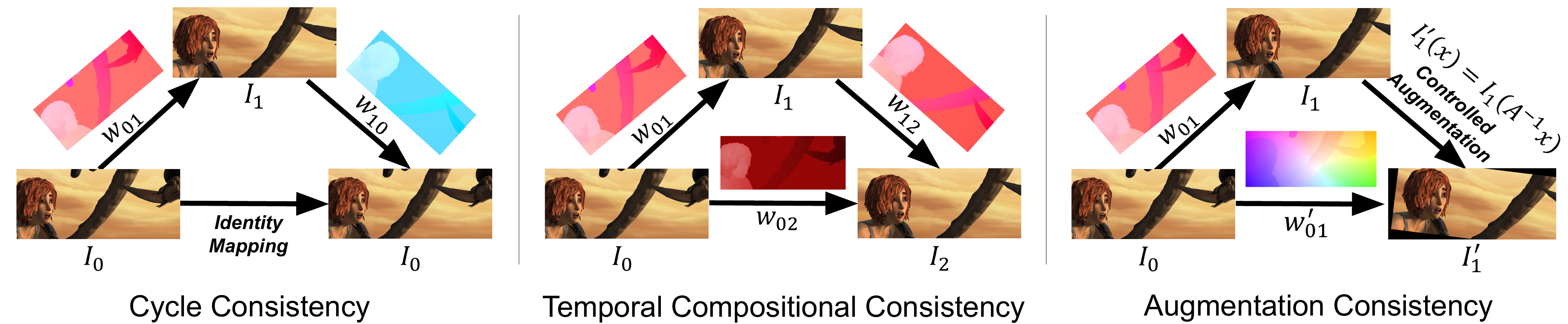}
\caption{\small 
{\bf Triangular consistency as a universal compositional principle.}
Optical flow fields compose. Triangular consistency enforces that the flow between two frames agrees with the composition of intermediate flows. 
Cycle consistency arises as a special case when the composition returns to the same frame (\emph{left}). 
The same principle naturally generalizes to video through temporal chaining across multiple frames (\emph{middle}), and further extends to synthetic transformations, enabling controlled data augmentation for optical flow without additional annotations (\emph{right}). 
}
\label{fig:triangular}
\end{figure*}

\noindent\textbf{Composition, chaining, and long-range correspondence.} Composition of displacement fields has long been exploited when moving beyond two-frame estimation~\cite{zhou2015flowweb,lao2017minimum,wang2019learning,lao2024diffeomorphic}. The theoretical basis of such compositional constraints is closely related to diffeomorphic warping theory~\cite{yang2014shape, yang2015self,sundaramoorthi2018variational}. In co-visible regions, image motion can be modeled as a smooth invertible warp, and diffeomorphisms are closed under composition; hence, the consistency of composed optical flows follows from the geometry of the underlying transformation space rather than from a heuristic temporal prior. On the other hand, layered motion models provide a principled view of why such warping may fail at occlusion boundaries and how multiple motion modes coexist~\cite{sun2012layered,lao2018extending,lao2021flow}, and recent benchmarks further emphasize multi-layer and non-Lambertian challenges~\cite{wen2024layeredflow}. In parallel, recent advances in dense tracking revisit optical flow as a building block for long-term trajectories through chaining and refinement~\cite{cho2024flowtrack,neoral2024mft,le2024dense,wang2023tracking,doersch2023tapir,karaev2024cotracker}. These works primarily employ flow composition to initialize correspondence. Our contribution differs in scope and objective. We formulate composition as an explicit supervision signal through the simplest non-trivial compositional structure, a \emph{triangle} (Fig.~\ref{fig:triangular}), and evaluate it across supervised, unsupervised, and adaptation settings. By treating composition as a universal learning constraint rather than solely as a tracking strategy, triangular consistency introduces a principled supervision mechanism that has not been explicitly incorporated into the optical flow training pipeline.

\section{Method}\label{sec:method}

\subsection{Formalization}
\label{sec:formalization}

Let $I_t : \Omega \rightarrow \mathbb{R}^k$ ($k=3$ for RGB) denote an image at time $t$, where $\Omega \subset \mathbb{R}^2$ is the image domain. Let $v_{t,t+1} : \Omega \rightarrow \mathbb{R}^2$ denote the optical flow from frame $t$ to $t+1$, and define the corresponding warp
$
w_{t,t+1}(x) = x + v_{t,t+1}(x).
$
Similarly, for any pair $(t,s)$ we write $w_{t,s}(x) = x + v_{t,s}(x)$.

\noindent\textbf{Compositional Structure.}
As in Sec.~\ref{sec:intro}, we view displacement fields as non-rigid coordinate transformations and therefore compose. Given three frames $I_t, I_{t+1}, I_{t+2}$ from the same scene, the composed warp from $t$ to $t+2$ is
\begin{equation}
\widetilde{w}_{t,t+2}(x)
= w_{t+1,t+2}\big(w_{t,t+1}(x)\big).
\label{eq:composition}
\end{equation}
In displacement form (i.e. optical flow vectors) this becomes
\begin{equation}
\widetilde{v}_{t,t+2}(x)
= v_{t,t+1}(x)
+ v_{t+1,t+2}\big(x + v_{t,t+1}(x)\big).
\label{eq:flow_composition}
\end{equation}

Triangular consistency requires that this composed flow agree with the directly estimated flow: $v_{t,t+2}(x)
\approx \widetilde{v}_{t,t+2}(x)$, which expresses the simplest non-trivial compositional relation among three frames. As shown in Fig.~\ref{fig:triangular}, cycle consistency is a special case where the composition returns to the same frame, while temporal chaining and controlled augmentation arise from the same rule.

\noindent\textbf{Triangular Consistency Loss.}
We thus define the triangular residual at pixel $x$ as
$
r_{t,t+1,t+2}(x)
=
v_{t,t+2}(x)
-
\widetilde{v}_{t,t+2}(x).
$
To mitigate the effect of outliers, we penalize it with a robust norm $\rho(\cdot)$, yielding triangular consistency loss
\begin{equation}
\mathcal{L}_{\mathrm{tri}}
=
\sum_{x \in \Omega}
M_{t,t+1,t+2}(x)\,
\rho\!\left( \| r_{t,t+1,t+2}(x) \|_2 \right),
\label{eq:tri_loss}
\end{equation}
where $M_{t,t+1,t+2}(x) \in [0,1]$ is a validity mask discussed below.

\noindent\textbf{Occlusion.} In 2D images, pixels may become occluded or disoccluded between frames. In such regions, the mapping $w_{t,t+2}$ is not the composition of visible correspondences unless a layered motion model is available~\cite{jackson2008dynamic,sun2012layered,lao2018extending}. Since typical optical flow methods do not explicitly maintain a multi-layer representation, compositional consistency is violated at occlusion boundaries.

To address this, we construct $M_{t,t+1,t+2}(x)$ using forward-backward consistency checks. Regions that violate such consistency constraints are down-weighted. This ensures that triangular consistency is enforced primarily on regions where valid motion correspondences can be established across frames.

\subsection{Learning Objective}
\label{sec:learning_objective}

\noindent\textbf{Cycle Consistency.}
When $(i,j,k)=(t,t+1,t)$, triangular consistency reduces to forward-backward cycle consistency:
$
v_{t,t+1}(x) + v_{t+1,t}(x + v_{t,t+1}(x)) \approx 0.
$
The corresponding loss is
$
\mathcal{L}_{\mathrm{cyc}}
=
\sum_x
\rho\!\left(
\| v_{t,t+1}(x) + v_{t+1,t}(x + v_{t,t+1}(x)) \|_2
\right).
$

\noindent\textbf{Temporal Compositional Consistency.}
For three distinct frames $(i,j,k)=(t,t+1,t+2)$, we enforce
$
v_{t,t+2}(x)
=
v_{t,t+1}(x)
+
v_{t+1,t+2}(x + v_{t,t+1}(x)),
$
leading to
$
\mathcal{L}_{\mathrm{temp}}
=
\sum_x
\rho\!\left(
\| v_{t,t+2}(x) - \widetilde{v}_{t,t+2}(x) \|_2
\right).
$

\noindent\textbf{Augmentation Consistency.}
Let $A$ be a known affine transformation. Given $I_1' = A(I_1)$, the induced analytical flow from $I_1$ to $I_1'$ is $v_{1,1'}(x)=A(x)-x$. Triangular consistency enforces
$
v_{0,1'}(x)
=
v_{0,1}(x)
+
v_{1,1'}(x + v_{0,1}(x)).
$
The loss is
$
\mathcal{L}_{\mathrm{aug}}
=
\sum_x
\rho\!\left(
\| v_{0,1'}(x)
-
v_{0,1}(x)
-
v_{1,1'}(x + v_{0,1}(x)) \|_2
\right).
$
\smallskip

\noindent\textbf{Complete Objective.}
Let $\mathcal{L}_{\mathrm{base}}$ denote the baseline loss (supervised loss, photometric reconstruction, flow smoothness, etc.). The overall objective becomes
\begin{equation}
\mathcal{L}_{\mathrm{total}}
=
\mathcal{L}_{\mathrm{base}}
+
\lambda_{\mathrm{cyc}} \mathcal{L}_{\mathrm{cyc}}
+
\lambda_{\mathrm{temp}} \mathcal{L}_{\mathrm{temp}}
+
\lambda_{\mathrm{aug}} \mathcal{L}_{\mathrm{aug}}.
\label{eq:total_loss}
\end{equation}

As such, the training objective Eq.~\eqref{eq:total_loss} is architecture-agnostic and can be directly integrated into existing optical flow models without modification, functioning as a plug-and-play supervision component.

\subsection{Implementations}
\label{sec:implementation}
\noindent\textbf{Flow Composition.}
In Eq.~\eqref{eq:flow_composition}, the term $v_{t+1,t+2}\big(x + v_{t,t+1}(x)\big)$ requires evaluating $v_{t+1,t+2}$ at displaced coordinates. This is implemented by bilinear interpolation on a coordinate grid. Coordinates that fall outside image bounds are treated as invalid and excluded from the loss. Importantly, we do not warp images. Instead, the composition is applied directly to the flow field. As a result, the supervision depends purely on the geometric relation between coordinates and is independent of color, texture, or image-feature consistency.

\noindent\textbf{Controlled Augmentation.}
For augmentation consistency, we sample a transformation consisting of translation $t=(t_x,t_y)^\top$, rotation $\theta$, and scale $s$. Given the image center $c=(c_x,c_y)^\top$, the forward transform from $I_1$ to $I'_1$ is
\begin{equation}
A(x)=R_s(x-c)+c+t,
\qquad
R_s=s
\begin{pmatrix}
\cos\theta & -\sin\theta\\
\sin\theta & \cos\theta
\end{pmatrix}.
\end{equation}
For resampling, we use the corresponding inverse map $A^{-1}(x)=Mx+b$, where
\begin{equation}
M=R_s^{-1}
=
\frac{1}{s}
\begin{pmatrix}
\cos\theta & \sin\theta\\
-\sin\theta & \cos\theta
\end{pmatrix},
\qquad
b=c-M(c+t).
\end{equation}
This gives an exact inverse mapping, and $I'_1$ can be sampled from $I_1$ accordingly.

Now let $v_{01}$ be the optical flow from $I_0$ to $I_1$, either from the ground truth or model prediction. Define base pixel coordinates $x$ and transported coordinates $y = x + v_{01}(x)$. The augmented target position is

\begin{equation}
y' = A(y).
\end{equation}

The induced flow from $I_0$ to the augmented frame $I_1'$ is then

\begin{equation}
v_{01}'(x) = y' - x. 
\label{eq:augmentation}
\end{equation}

Eq.~\eqref{eq:augmentation} is entirely analytical and computed in closed form. The augmented flow $v_{01}'$ is thus obtained through direct coordinate transformation and does not require any image resampling or grid-based interpolation. Consequently, the supervision signal is free from interpolation artifacts regardless of the magnitude or complexity of the sampled transformation.

Importantly, although the augmented image may contain invalid regions due to pixels mapping outside the image domain, the induced optical flow field remains well-defined for \emph{every} pixel in the source frame. The benefit of this property is non-trivial for supervising optical flow: the augmented ground-truth flow preserves all the spatial regularities and smoothness of the original displacement field without invalid pixels at image boundaries or artifacts caused by resampling. As a result, triangular augmentation remains free from artifacts even under aggressive transformations.

\noindent\textbf{Occlusion Filtering.}
Occlusion violates cycle consistency and temporal compositional consistency. We therefore need to explicitly exclude occluded regions in the training loss. Specifically, we compose forward and backward flows, and a pixel is considered valid if it lands close to its original position after forward and then backward mapping. Pixels that violate this consistency are down-weighted through an occlusion mask derived from the residual. Importantly, this procedure evaluates the consistency of coordinate mappings rather than color similarity via reprojection~\cite{ren2017unsupervised}. In practice, we compute this compositional residual directly from the predicted flows, and weight the loss using a robust norm (Eq.~\eqref{eq:tri_loss}), so that unreliable regions are softly suppressed. 

\noindent\textbf{Speed.}
The proposed constraint introduces negligible computational overhead. At $386\times496$, the interpolation required for flow composition takes approximately $0.00030$ seconds, while affine augmentation for both the image and optical flow requires $0.00073$ seconds on an NVIDIA RTX Pro 6000 Blackwell GPU, which is negligible in practice. For example, in the setting of Sec.~\ref{sec:adapt}, when training a RAFT-Large~\cite{teed2020raft} model, the compositional operations and loss evaluation in total account for $0.12\%$ of the total training wall-clock time, indicating that triangular consistency can be incorporated into existing training pipelines with essentially no measurable slowdown.


\section{Experiments}
\label{sec:experiments}
To isolate the effect of the proposed compositional constraint, we select two widely used yet relatively minimal optical flow frameworks as baselines. For unsupervised learning, we adopt \textbf{ARFlow}~\cite{liu2020learning}.
For self-supervised adaptation and supervised learning, we use \textbf{RAFT}~\cite{teed2020raft}.
These choices allow us to attribute performance changes directly to triangular consistency without auxiliary engineering components.
Many subsequent methods, including SMURF~\cite{Stone_2021_CVPR}, largely combine ARFlow-style self-supervised objectives with RAFT-style architectures. Crucially, our method introduces only additional supervision and does not alter the estimator or the original training loss. Thus, it is compatible with existing and future optical flow methods, and improvements observed in our experiments are expected to transfer to subsequent methods.

We test across synthetic and real-world datasets:
\textbf{FlyingChairs}~\cite{dosovitskiy2015flownet} and \textbf{FlyingThings3D}~\cite{mayer2016large} provide large synthetic motion with diverse displacements and occlusions;
\textbf{MPI-Sintel}~\cite{butler2012naturalistic} offers complex non-rigid motion, and its \textit{Final} setting introduces effects such as motion blur and atmospheric distortions;
\textbf{KITTI}~\cite{geiger2012we} is a real driving benchmark with ground truth derived from LiDAR/3D reconstruction. We also report zero-shot transfer results on \textbf{HD1K}~\cite{kondermann2016hci} (high-resolution driving sequences) and \textbf{Middlebury}~\cite{baker2011database} (real scenes with small-to-medium motions).
We report standard endpoint error (EPE). For KITTI, we report Fl-all, the percentage of optical flow outliers over all ground-truth pixels. A pixel is counted as an outlier when its endpoint error is at least 3 pixels and at least 5\% of the ground-truth flow magnitude~\cite{menze2015object}.

\begin{figure*}[t]
\centering
\includegraphics[width=0.8\textwidth]{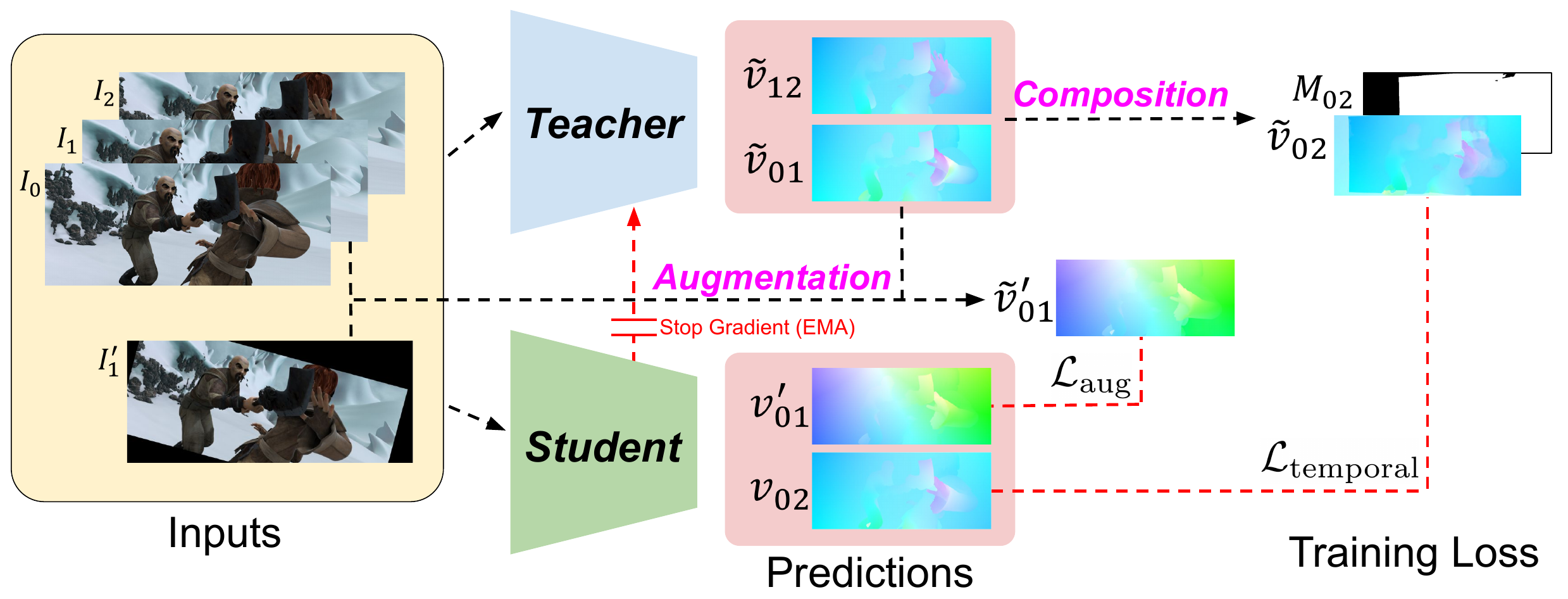}
\caption{\small 
\textbf{Self-supervised adaptation with triangular consistency.}
Given input frames $(I_0,I_1,I_2)$, the teacher network predicts flows $\tilde{v}_{01}$ and $\tilde{v}_{12}$, which are composed to produce the reference flow $\tilde{v}_{02}$. The student network predicts $v_{02}$ and is trained by enforcing consistency with this composed reference. In parallel, a random affine transformation generates an augmented frame $\tilde{I}_1$, inducing the analytically defined flow $\tilde{v}_{01}'$. The student prediction $v_{01}'$ is required to match this reference. The teacher parameters are updated using an exponential moving average (EMA) of the student.
}
\label{fig:distill}
\end{figure*}

\subsection{Self-Supervised Adaptation}
\label{sec:adapt}

We first test how triangular consistency \emph{alone} can improve an optical flow model. Our goal is to push the limit of using \emph{only} consistency as supervision, without any auxiliary signals (e.g., photometric reconstruction or smoothness), even if they do not require labels and may result in additional improvement. To this end, we adapt pre-trained optical flow models to new domains without using labeled data in a few-shot setting: a \emph{single} epoch of self-supervised adaptation, after which the adapted model is evaluated on frames that were \emph{not used} during the adaptation stage, resembling a practical test-time adaptation scenario.

\noindent\textbf{Method.} We start from RAFT models pre-trained on FlyingChairs or FlyingThings3D and adapt them to Sintel using neither labels nor photometric or smoothness losses: the only training signal is triangular consistency. Note that we perform adaptation on Sintel's unlabeled test split and evaluate on its labeled training split.\footnote{Sintel's official evaluation server does not permit repeated evaluation on the test split. We therefore use the available labels from the training split only for evaluation. This experiment is intended as a controlled test of triangular consistency; in practice, the same adaptation can be combined with other forms of self-supervision.} This yields a lightweight, test-time-style calibration setting: with batch size 12, one epoch corresponds to 45 iterations and takes 84 seconds on an NVIDIA RTX Pro 6000 Blackwell GPU. Notably, to isolate the effect of the proposed constraint, we freeze normalization statistics during adaptation, so that the model performs no running-stat updates during training (verified by a sanity check). This practice prevents misattributing spurious gains from trivially recalibrating batch statistics to the target data domain.

To stabilize the adaptation process, we adopt a teacher-student self-distillation scheme~\cite{he2020momentum}. The loss is back-propagated only through the student model, while the teacher model is updated using an exponential moving average (EMA) of the student parameters. This scheme is summarized in Fig.~\ref{fig:distill}. The teacher predicts the shorter-range flows $\tilde{v}_{01}$ and $\tilde{v}_{12}$, while the student predicts the longer-range flow $v_{02}$. We then penalize the discrepancy between the student's direct prediction $v_{02}$ and the composed teacher prediction $\widetilde{v}_{02}$. This effectively chains two smaller-displacement, typically more accurate, flows to supervise a larger-displacement flow. In parallel, we enforce augmentation consistency similarly. A random affine transformation is sampled and applied to frame $I_1$ to produce the augmented frame $\tilde{I}_1$, which also induces flow $\tilde{v}'_{01}$. The loss is computed between the student prediction $v_{01}'$ and the reference flow $\tilde{v}'_{01}$.

\begin{wraptable}{r}{0.54\linewidth}

\centering
\caption{\small 
\textbf{Self-supervised adaptation by triangular consistency.}
Under extremely constrained settings: one epoch (45 iterations) using unlabeled data and \emph{only} triangular consistency, accuracy improves significantly.}
\label{tab:adapt}\small
{\setlength{\tabcolsep}{2pt} 
\begin{tabular}{lcccc}
\toprule
Pre-training & Method & Clean & Final\\
\midrule
\multirow{3}{*}{FlyingChairs}
 & Pre-trained & 2.54 & 4.67\\
 & + consistency & \textbf{2.08} & \textbf{3.95}\\
 & Improvement & 18.1\% & 15.4\%\\
\midrule
\multirow{3}{*}{FlyingThings3D}
 & Pre-trained & 2.01 & 3.41 \\
 & + consistency & \textbf{1.73} & \textbf{3.13}\\
 & Improvement & 13.9\% & 8.2\%\\
\bottomrule
\end{tabular}}
\end{wraptable}

\noindent\textbf{Results.}
Tab.~\ref{tab:adapt} demonstrates that consistency alone already leads to substantial improvements in optical flow. Even with an extremely small adaptation budget, the pre-trained models consistently improve across Sintel benchmarks by up to 18.1\%. These gains are particularly notable given the constrained setting: adaptation is performed using only unlabeled data and relies solely on the proposed consistency objectives, without photometric losses, smoothness losses, or ground-truth labels. Among the proposed components, temporal consistency provides the dominant adaptation signal, improving the error from 2.01 to 1.77 and from 2.54 to 2.18. The augmentation objective further improves these results from 1.77 to 1.73 and from 2.18 to 2.08. In contrast, augmentation-only and cycle-only variants are ineffective, likely because temporal consistency supervises longer-range flow using two shorter, typically more reliable flows, whereas the other terms do not provide a comparable adaptation signal. The fact that such a minimal signal can reliably improve pre-trained models suggests that triangular consistency provides a strong source of supervision for domain adaptation. This is particularly valuable for active learning~\cite{Yuan2022OpticalFT}, where high-quality labeled data are scarce and costly to obtain.

\subsection{Unsupervised Training}

Encouraged by the improvements in the self-supervised adaptation, we now integrate all three forms of triangular consistency into a complete unsupervised optical flow training pipeline.

\noindent\textbf{Method.}
We build upon ARFlow~\cite{liu2020learning}, a widely used unsupervised optical flow framework. ARFlow natively supports loading three consecutive frames during training, allowing triangular consistency to be incorporated without modifying the data loading pipeline. ARFlow primarily relies on photometric reconstruction together with occlusion-aware bidirectional consistency and symmetric augmentation consistency between transformed image pairs as supervision signals. Note that our augmentation consistency differs from ARFlow, since it only augments the target image and is therefore asymmetric. Unlike the self-supervised adaptation experiment, which adapts pre-trained models using unlabeled data, here we train models from scratch using the unlabeled Sintel and KITTI training sets. All training settings and hyperparameters, other than our proposed training loss, follow the publicly available code of ARFlow.

\begin{figure*}[t]
\centering
\includegraphics[width=\textwidth]{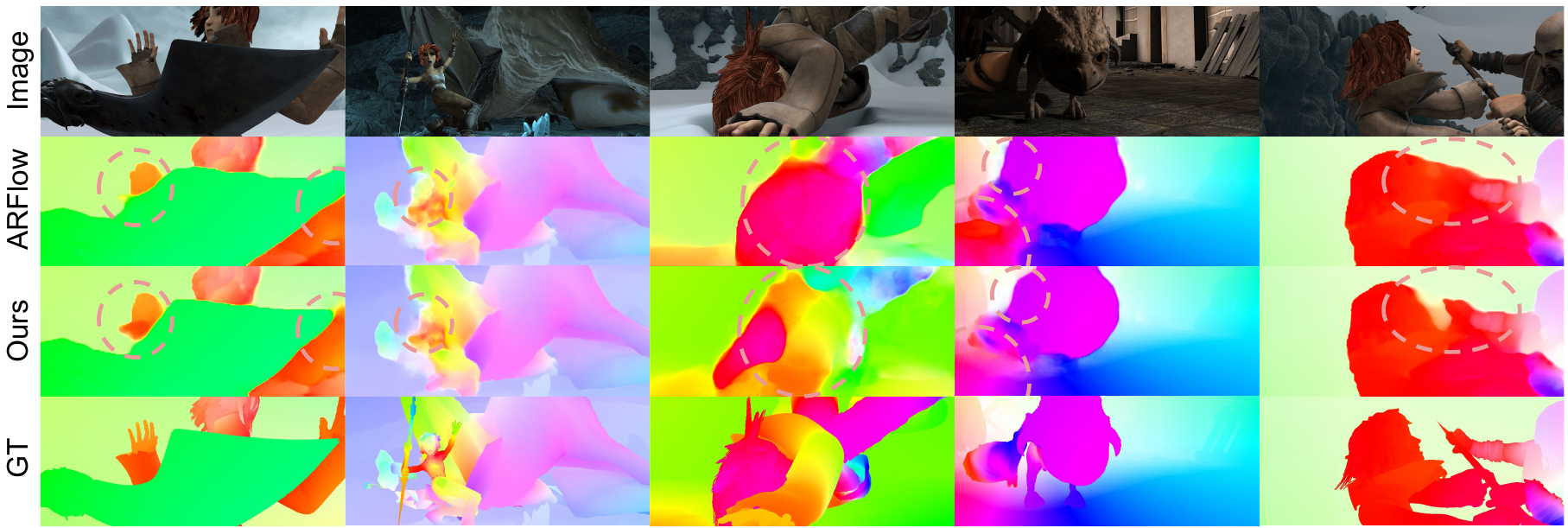}
\caption{\small 
\textbf{Qualitative comparison on MPI-Sintel.}
Regions highlighted by dashed circles illustrate typical improvements introduced by triangular consistency. Our method produces motion fields that better align with the object boundaries and motion continuity of the scene, reflecting the geometric consistency enforced during training.
}
\label{fig:comparison}
\end{figure*}

\begin{table*}[t]
\centering
\caption{\small
\textbf{Unsupervised learning with triangular consistency.}
When training on Sintel, triangular consistency improves both training/test accuracy and cross-dataset performance. When training on KITTI, the training/test error remains nearly unchanged, but zero-shot transfer improves noticeably.}
\label{tab:unsup_results}
\resizebox{\linewidth}{!}{
{\setlength{\tabcolsep}{5pt}
\begin{tabular}{l |l| cc |c |c|c}
\toprule
& & \multicolumn{2}{c|}{Sintel (train/test)} & KITTI (train/test) & HD1K & Middlebury \\
Source & Method & Clean EPE & Final EPE & Fl-all (\%) & EPE & EPE \\
\midrule
\multirow{3}{*}{Sintel}
& ARFlow & 2.79 / 4.78 & 3.73 / 5.89 & - & 1.40 & 0.35 \\
& + ours & \textbf{2.58 / 4.48} & \textbf{3.49 / 5.86} & - & \textbf{1.24} & \textbf{0.33} \\
& Improvement & 7.5\% / 6.3\% & 6.4\% / 0.5\% & - & 11.4\% & 5.7\% \\
\midrule
\multirow{3}{*}{KITTI}
& ARFlow & - & - & \textbf{9.87} / 11.80 & 2.32 & 0.55 \\
& + ours & - & - & 9.94 / \textbf{11.44} & \textbf{1.99} & \textbf{0.54} \\
& Improvement & - & - & -0.7\% / 3.1\% & 14.2\% & 1.8\% \\
\bottomrule
\end{tabular}}}
\end{table*}

\noindent\textbf{Results.} Fig.~\ref{fig:comparison} presents qualitative comparisons on several challenging Sintel sequences. Because triangular consistency constrains the geometry of motion correspondences, its effect is visible in the structural coherence of the predicted flow fields. In particular, motion boundaries and large coherent regions exhibit improved alignment with the ground truth. The highlighted regions illustrate typical cases where enforcing consistency across composed flows leads to more stable motion estimates, especially for articulated objects and regions undergoing complex motion.  Tab.~\ref{tab:unsup_results} shows that adding triangular consistency improves unsupervised learning in both \emph{in-domain} accuracy and \emph{cross-dataset} transfer. When training on Sintel, we reduce EPE on both Clean/Final and also improve zero-shot evaluation on HD1K and Middlebury, indicating that compositional and analytic-augmentation constraints provide supervision complementary to photometric reconstruction and bidirectional consistency.

When training on KITTI, metrics on the training set saturate: there is little improvement on KITTI training itself, yet test accuracy still improves by 3.1\%. This already suggests that the added constraint primarily improves \emph{generalization}, especially when the training set data distribution is narrow. The benefit becomes more evident when evaluating zero-shot transfer. Models trained with triangular consistency achieve substantially better performance on HD1K, reducing EPE by 14.2\%. Notably, the same trend is observed in the supervised setting, where triangular consistency introduces motion variations that improve test accuracy and cross-dataset generalization even when the training-set accuracy itself is slightly reduced, reflecting less overfitting.

\begin{table}[t]
\centering\small
\caption{\textbf{Ablation of triangular consistency losses and loss weights.}
We report Sintel Clean and Final EPE under unsupervised ARFlow training; lower is better.}
\label{tab:ablation_weights}
\begin{tabular}{lcc}
\toprule
Configuration & Clean & Final \\
\midrule
Baseline (ARFlow) & 2.79 & 3.73 \\
\midrule
+ Aug ($\lambda_{\mathrm{aug}}=0.003$) & 2.72 & 3.70 \\
+ Aug ($\lambda_{\mathrm{aug}}=0.01$) & 2.60 & 3.56 \\
\midrule
+ Temp ($\lambda_{\mathrm{temp}}=0.003$) & 2.65 & 3.68 \\
+ Temp ($\lambda_{\mathrm{temp}}=0.01$) & 2.67 & 3.69 \\
\midrule
+ Cyc ($\lambda_{\mathrm{cyc}}=0.003$) & 2.76 & 3.77 \\
+ Cyc ($\lambda_{\mathrm{cyc}}=0.005$) & 2.69 & 3.70 \\
+ Cyc ($\lambda_{\mathrm{cyc}}=0.01$) & 2.68 & 3.69 \\
\midrule
+ Temp + Cyc ($\lambda_{\mathrm{temp}}=0.003$, $\lambda_{\mathrm{cyc}}=0.005$) & 2.67 & 3.68 \\
+ Aug + Temp ($\lambda_{\mathrm{aug}}=0.01$, $\lambda_{\mathrm{temp}}=0.003$) & 2.63 & 3.59 \\
+ Aug + Cyc ($\lambda_{\mathrm{aug}}=0.01$, $\lambda_{\mathrm{cyc}}=0.005$) & 2.61 & 3.57 \\
\midrule
+ Aug + Temp + Cyc ($\lambda_{\mathrm{aug}}=0.02$, $\lambda_{\mathrm{temp}}=0.003$, $\lambda_{\mathrm{cyc}}=0.005$) & 2.68 & 3.53 \\

+ Aug + Temp + Cyc ($\lambda_{\mathrm{aug}}=0.01$, $\lambda_{\mathrm{temp}}=0.003$, $\lambda_{\mathrm{cyc}}=0.005$) & \textbf{2.58} & \textbf{3.49} \\
\bottomrule
\end{tabular}
\end{table}

\noindent\textbf{Ablations.}
Tab.~\ref{tab:ablation_weights} evaluates the contribution of the three types of triangular consistency. Augmentation consistency gives the largest improvement, reducing the ARFlow baseline from 2.79/3.73 to 2.60/3.56 when $\lambda_{\mathrm{aug}}=0.01$. Temporal compositional consistency also improves over the baseline, but its gains are smaller and remain similar across the tested weights. Cycle consistency is less stable in isolation: moderate weights improve Clean performance, while Final remains close to the baseline. The combined variants follow the same pattern. Configurations containing augmentation generally outperform temporal and cycle consistency combinations, and the full combination of the three training objectives gives the best overall results, achieving the lowest Clean EPE with $(\lambda_{\mathrm{aug}},\lambda_{\mathrm{temp}},\lambda_{\mathrm{cyc}})=(0.01,0.003,0.005)$. This suggests that augmentation consistency is the dominant signal, unlike in self-supervised adaptation where temporal consistency is more effective. A possible reason is that training from scratch and adapting a pre-trained model have different optimization dynamics.

\subsection{Supervised Training}

Finally, we evaluate triangular consistency under a supervised setting. In this setting, we strictly follow the original RAFT training pipeline~\cite{teed2020raft} and introduce triangular consistency only as a controlled data augmentation mechanism.

\noindent\textbf{Method.}
We apply random affine transformations to the target image. Using the analytic formulation described in Sec.~\ref{sec:implementation}, the corresponding pseudo-ground-truth flow is computed accordingly. The model is then trained on the augmented image pair and pseudo-ground-truth. This augmentation effectively preserves exact supervision since the pseudo-ground-truth is computed analytically. Importantly, this is the only modification to the RAFT training pipeline and introduces negligible computational overhead. All other training settings, including model and hyperparameters, remain unchanged.

\begin{table*}[t]
\centering
\caption{\small 
\textbf{Triangular consistency as data augmentation for supervised training.}
We follow the RAFT training pipeline and introduce triangular consistency only through controlled data augmentation. This slightly improves test accuracy and further improves zero-shot transfer across datasets.}
\label{tab:supervised}
\resizebox{\linewidth}{!}{
{\setlength{\tabcolsep}{6pt} 
\begin{tabular}{l |l| cc |c |c|c}
\toprule
& & \multicolumn{2}{c|}{Sintel (train/test)} & KITTI (train/test) & HD1K & Middlebury \\
Source & Method & Clean EPE & Final EPE & Fl-all (\%) & EPE & EPE \\
\midrule
\multirow{3}{*}{Sintel}
& RAFT & 0.76 / 2.08 & 1.22 / \textbf{3.41} & - & 0.67 & 0.29 \\
& + ours & \textbf{0.71 / 2.02} & \textbf{1.16} / 3.44 & - & \textbf{0.66} & \textbf{0.27} \\
& Improvement & 6.6\% / 2.9\% & 4.9\% / -0.9\% & - & 1.5\% & 6.9\% \\
\midrule
\multirow{3}{*}{KITTI}
& RAFT & - & - & \textbf{1.53} / 5.27 & 1.08 & 0.69 \\
& + ours & - & - & 1.56 / \textbf{5.02} & \textbf{0.83} & \textbf{0.56} \\
& Improvement & - & - & -2.0\% / 4.7\% & 23.1\% & 18.8\% \\
\bottomrule
\end{tabular}}}
\end{table*}

\noindent\textbf{Results.}
Tab.~\ref{tab:supervised} shows that triangular consistency consistently improves optical flow under supervised training. We note that Sintel contains large synthetic motions, e.g., action and combat sequences. This induces a bias when transferring to datasets such as Middlebury, which capture motions from natural interactions. By using controlled affine transformations to introduce additional camera-like motion patterns, triangular consistency serves as a regularizer that enables RAFT trained on Sintel to generalize to Middlebury. Hence, despite Sintel having high-fidelity supervision, the proposed triangular consistency still improves cross-dataset generalization by 6.9\%. Interestingly, our method does not improve accuracy on the Sintel Final pass, which mainly introduces photometric variations rather than geometric ones. This lack of improvement is consistent with the source of our gains: triangular consistency improves geometric motion generalization, but is not designed to address appearance changes.

The training-set bias is even more pronounced when training on KITTI and testing on HD1K and Middlebury. Since KITTI is an outdoor driving dataset, the motion patterns are largely constrained: forward ego-motion dominates, and viewpoints, motion directions, and displacement magnitudes follow a relatively predictable pattern determined by vehicle dynamics and frame rate. In contrast, Middlebury contains significant motions of people interacting with objects, while the camera motion remains conservative. Training with triangular consistency naturally introduces a much larger set of motion patterns to KITTI. This is reflected in reduced overfitting to the training set and moderate (4.7\%) improvement on the KITTI test set, but more substantial improvement when transferring across datasets: 23.1\% on HD1K and 18.8\% on Middlebury.

This experiment highlights the effectiveness of such a data augmentation mechanism for supervised optical flow. To the best of our knowledge, existing augmentation strategies preserve correspondence by applying identical transformations to both images in a pair~\cite{liu2020learning,teed2020raft,Stone_2021_CVPR}. In contrast, we transform only the target image and analytically update the ground-truth. This enables the model to observe a broader set of motion patterns while maintaining exact supervision. Note that, in supervised training, if synthetic data could be generated indefinitely with user-specified motion patterns, such augmentation would be less necessary. In practice, however, synthetic datasets are often released without the simulator, while real-world datasets cannot be resynthesized at scale. In this sense, our method acts as a simulator. The results demonstrate the benefit of such augmentation, particularly under zero-shot transfer when motion statistics are highly constrained in the source dataset.

\section{Discussion and Conclusion}
\subsection{Why was such a simple constraint overlooked?}

At first glance, triangular consistency may appear almost self-evident. The compositional nature of displacement fields has been recognized and exploited in tasks involving long-range correspondence~\cite{lao2018extending, cho2024flowtrack, le2024dense}. We were therefore somewhat surprised that such a straightforward geometric relation has not been systematically used for training optical flow. One likely reason is historical: optical flow estimation has primarily been viewed as a pairwise problem~\cite{lucas1981iterative,horn1981determining}. Although multi-frame formulations were explored in classical optimization-based methods, they typically required solving large coupled systems over many frames~\cite{irani1999multi,garg2010dense}, making them computationally demanding. Modern learning-based optical flow models~\cite{dosovitskiy2015flownet,sun2018pwc,teed2020raft} inherit this pairwise design, and most supervision signals are likewise defined only for pairs of frames.

There are also practical reasons. Existing optical flow training pipelines apply identical geometric transformations to both images to preserve correspondences~\cite{teed2020raft,liu2020learning,Stone_2021_CVPR}. In contrast, we augment only the target image and update the ground-truth flow analytically, producing a broader family of motion patterns. Moreover, naively composing flows or applying geometric augmentation often introduces interpolation artifacts or invalid correspondences, particularly near occlusion boundaries and image borders. Our affine formulation alleviates this issue since induced flow can be computed in closed form, even when the transformed image extends beyond the image domain. This avoids interpolation artifacts and allows triangular consistency to be implemented as a stable training signal. Conceptually, this strategy is similar to AugUndo~\cite{wu2024augundo}, which introduces controlled geometric perturbations to generate additional supervision for depth prediction. As such, we foresee this work extending to other data modalities, e.g., medical images \cite{zhang2024adaptive,zhang2024heteroscedastic}, and geometric tasks, e.g., monocular depth estimation \cite{fei2019geo,wong2019bilateral} and completion \cite{liu2022monitored,park2026orcas,wong2020unsupervised,wong2021unsupervised}. In retrospect, the idea may appear simple, but our experiments show that incorporating this geometric constraint leads to consistent improvements across multiple training regimes.

\subsection{Constraints for Optical Flow}\label{sec:geometric}

Most geometric constraints in optical flow are relatively local. A common example is spatial smoothness, which assumes neighboring pixels move coherently~\cite{horn1981determining,bailer2015flow,chen2016full,revaud2015epicflow,jiang2021learning}. Some attempts also employ 3D priors as auxiliary supervision signals~\cite{zou2018df, jiao2021effiscene, dong2023rethinking,poggi2025flowseek}. Another widely used constraint is forward-backward \emph{symmetry}~\cite{janai2018unsupervised,liu2020learning, shi2023videoflow}, written as $v_{10} \approx -v_{12}$, which differs from the forward-backward \emph{compositional} consistency we study, written as $w_{10}(w_{01}) \approx \text{Id}$. While effective, it stems from a linear approximation of the trajectory under a temporal-smoothness assumption. Compared to these priors, triangular consistency arises from a more fundamental property: optical flow represents a mapping between coordinate systems, and such mappings compose by construction. Importantly, this relation holds regardless of the specific trajectory taken by the point.

Our compositional constraints do not depend on image appearance and are therefore robust to illumination changes, shadows, reflections, or imaging noise. In this sense, triangular consistency additionally encourages coherent motion estimates across frames, which is evident in Fig.~\ref{fig:comparison}. Unlike spatial and temporal smoothness priors, which require users to specify a desired level of smoothness, our proposed constraint does not introduce a smoothness prior. Further, in this work, we focus only on the simplest non-trivial instance of such a compositional structure: a triangle. In principle, these transformations can be chained indefinitely over longer temporal sequences, potentially enabling stronger supervision signals. In future work, this may lead to a curriculum-learning pipeline, where the model is trained with progressively increasing correspondence ranges.

\subsection{Limitations}\label{sec:limitation}

Despite its generality, the effectiveness of triangular consistency depends on the motion statistics of the data. For example, we observe less improvement on the KITTI benchmark, which contains highly restrained motion patterns dominated by forward ego-motion in driving scenes. Consequently, viewpoints, motion directions, and displacement magnitudes follow relatively predictable patterns. In such environments, compositional constraints provide extra supervision, but the benefit is less visible when evaluation follows the original in-domain motion statistics.
Importantly, triangular consistency does not hurt performance on KITTI, while models trained on KITTI still demonstrate improvement when tested on HD1K and Middlebury. This observation suggests that geometric consistency constraints are most beneficial when the data contains complex and varied motion that cannot be captured by simple priors.

Another limitation arises from occlusion and multi-layer motion. The compositional relation assumes that a single-layer correspondence exists across frames. In real scenes, occlusions and independently moving objects may violate this assumption. While our implementation mitigates this issue through occlusion masking, incorporating explicit layered motion models or more advanced visibility reasoning may further improve robustness in such scenarios.

\subsection{Conclusion}
In this work, we revisit a simple yet fundamental property of motion: displacement fields agree under composition. From this intuition, we propose triangular consistency, a minimal compositional constraint that links optical flow estimation across three frames. It depends only on the geometry of motion and therefore remains agnostic to image appearance, network architecture, and supervision type.  The resulting method can be integrated into existing optical flow learning pipelines with negligible computational overhead. Despite its simplicity, the proposed mechanism consistently improves optical flow accuracy across unsupervised learning, self-supervised adaptation, and supervised training. In summary, it is fast to compute and functions as a plug-and-play supervision component compatible with existing methods. More broadly, this compositional view is not specific to optical flow: any correspondence problem arising from a temporally discretized continuous process can, in principle, be formulated through consistency under composition.

%
%

\section*{Acknowledgements}

This research was supported by Dong Lao's startup funds at LSU. Computational resources for model training were partially provided by LSU HPC.

We thank Ganesh Sundaramoorthi for bringing up, as early as 2015, the idea of solving long-range correspondence by composition when working on low-latency moving object detection. The composition rule developed at that time later translated into multiple works, which appeared at CVPR 2017, ECCV 2018, ICCV 2021, and CVPR 2024. The idea of controlled geometric augmentation originated from Alex Wong during the development of AugUndo for depth estimation. The initial concept of this work was partially developed at the UCLA Vision Lab during Dong Lao's and Alex Wong's appointments at UCLA.

We would like to give special thanks to the anonymous reviewers for encouraging and appreciating the honest, in-depth discussion of limitations presented in this paper. While demonstrating improvements is important, knowing when and why a method does not improve performance is equally important.

\bibliographystyle{splncs04}
\bibliography{main}
\end{document}